%% file: acl2023.tex
\pdfoutput=1
\documentclass[11pt]{article}

\usepackage[]{ACL2023}

\usepackage{times}
\usepackage{latexsym}

\usepackage[T1]{fontenc}
\usepackage[utf8]{inputenc}

\usepackage{microtype}

\usepackage{inconsolata}

\usepackage[utf8]{inputenc}
\usepackage{booktabs}
\usepackage{graphicx}
\usepackage{CJK}
\usepackage{multirow}
\usepackage{makecell}
\usepackage{color}
\usepackage{verbatim}
\usepackage{url}
\usepackage{enumitem}
\usepackage{amsmath}
\usepackage{diagbox}
\usepackage{flushend,cuted}
\usepackage{bbm}
\usepackage{xcolor}
\usepackage{xspace}
\usepackage{algorithm}
\usepackage{algpseudocode}
\usepackage{verbatim}
\usepackage{colortbl}
\usepackage{bbding}

\usepackage{listings}

\newcommand{\method}{INK\xspace}

\title{\method: Injecting $k$NN Knowledge in Nearest Neighbor Machine Translation}
\author{
    Wenhao Zhu$^{1}$ \text{,} \textbf{Jingjing Xu}$^{2}$ \text{,} \textbf{Shujian Huang}$^{1}$ \text{,} \textbf{Lingpeng Kong}$^{3}$\textbf{,} \textbf{Jiajun Chen}$^{1}$ \\
    $^{1}$ \text{National Key Laboratory for Novel Software Technology, Nanjing University} \\
    $^{2}$ \text{Shanghai AI Laboratory} $^{3}$ \text{The University of Hong Kong} \\
    \normalsize\texttt{zhuwh@smail.nju.edu.cn}, \normalsize\texttt{jingjingxu@pku.edu.cn} \\
    \normalsize\texttt{huangsj@nju.edu.cn}, 
    \normalsize\texttt{lpk@cs.hku.hk},
    \normalsize\texttt{chenjj@nju.edu.cn} \\
}

\begin{document}
\maketitle

\input{Latex/00_abstract.tex}

\input{Latex/01_introduction.tex}

\input{Latex/02_background.tex}

\input{Latex/03_method.tex}

\input{Latex/04_experiments.tex}

\input{Latex/05_analysis.tex}

\input{Latex/06_related_work.tex}

\input{Latex/07_conclusion.tex}

\input{Latex/08_limitation.tex}

% \normalem
\bibliography{custom}
% \bibliography{anthology}
\bibliographystyle{acl_natbib}

\input{Latex/09_appendix.tex}

\end{document}

%% file: Latex/00_abstract.tex
\begin{abstract}
Neural machine translation has achieved promising results on many translation tasks. However, previous studies have shown that neural models induce a non-smooth representation space,  which harms its generalization results. Recently, $k$NN-MT has provided an effective paradigm to smooth the prediction based on neighbor representations during inference. Despite promising results, $k$NN-MT usually requires large inference overhead. We propose an effective training framework \textbf{\method} to directly smooth the representation space via adjusting representations of $k$NN neighbors with a small number of new parameters. The new parameters are then used to refresh the whole representation datastore to get new $k$NN knowledge asynchronously. This loop keeps running until convergence. Experiments on four benchmark datasets show that \method achieves average gains of 1.99 COMET and 1.0 BLEU, outperforming the state-of-the-art $k$NN-MT system with $\textrm{0.02} \times$ memory space and 1.9$\times$ inference speedup\footnote{Code will be released at \url{https://github.com/OwenNJU/INK}}.
\end{abstract}

%% file: Latex/01_introduction.tex
\section{Introduction}
Neural machine translation (NMT) have achieved promising results in recent years \cite{vaswani2017attention, ng2019facebook, qian2021volctrans}. The target of NMT is to learn a generalized representation space to adapt to diverse scenarios. However, recent studies have shown that neural networks, such as BERT and GPT, induce non-smooth representation space, limiting the generalization abilities~\cite{gao2018representation, ethayarajh2019contextual, li2020sentence}. In NMT, we also observe a similar phenomenon in the learned representation space where low-frequency tokens disperse sparsely, even for a strong NMT model (More details are described in Section Experiments). Due to the sparsity, many ``holes'' could be formed. When it is used to translate examples from an unseen domain, the performance drops sharply~\cite{wang2022efficient, wang2022learning}

Recently, $k$-Nearest-Neighbor Machine Translation ($k$NN-MT)~\citep{khandelwal2021nearest} provides an effective solution to smooth predictions by equipping an NMT model with a \textit{key}-\textit{value} datastore.
For each entry, the \textit{value} is the target token and \textit{key} is the contextualized representation at the target position.  It requires a training set to record tokens and representations.
By aggregating nearest neighbors during inference, the NMT model can achieve decent translation results~\cite{khandelwal2021nearest, zheng2021adaptive, jiang2022towards}. Despite the success, $k$NN-MT also brings new issues with the increasing scale of training data. Retrieving neighbors from a large datastore~\cite{wang2022efficient} at each decoding step is time-consuming~\cite{martins2022efficient, meng2022fast}. Furthermore, once the datastore is constructed, representations can not be easily updated, limiting the performance ceiling of $k$NN-MT.

\begin{figure}[t]
    \centering
    \includegraphics[width=0.8\linewidth]{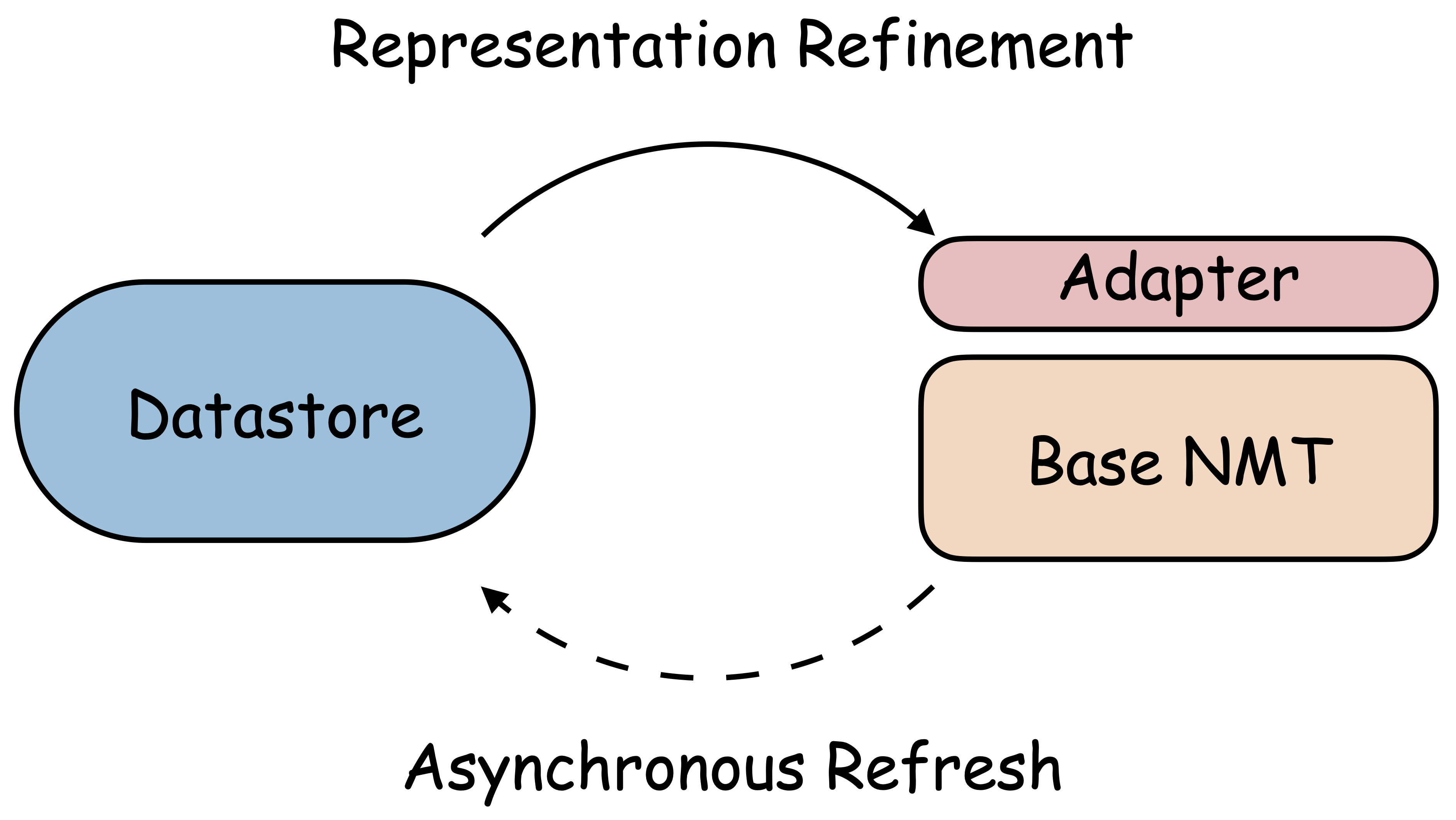}
    \caption{The overview of our training loop.  We refine the representation space of an NMT model according to the extracted $k$NN knowledge. The new parameters are then used to refresh the datastore to update $k$NN knowledge asynchronously.} 
    \label{fig:async}
\end{figure}

Given above strengths and weaknesses of $k$NN-MT, we propose to directly smooth the representation space with a small number of parameters.
In this paper, we propose a training framework \textbf{\method}, to iteratively refine the representation space with the help of extracted $k$NN knowledge (Fig. \ref{fig:async}).
Specifically, we adjust the representation distribution by aligning three kinds of representations with Kullback-Leibler (KL) divergence to train a small number of adaptation parameters. 
First, we align the contextualized representation and its target embedding to keep semantic meanings. 
Second, we align the contextualized representations of a target token and align the extracted $k$NN contextualized representations to address the sparsely dispersing problem.
After a training epoch, we refresh the datastore asynchronously with refined models to update $k$NN representations. During inference, we only load the off-the-shelf NMT model and tune adaptation parameters.

We conduct experiments on four benchmark datasets.
Experiment results show that our framework brings average gains of 1.99 COMET and 1.0 BLEU.
Compared with the state-of-the-art $k$NN-MT method (i.e. Robust $k$NN-MT; \citealt{jiang2022towards}), \method achieves better translation performance with 0.02$\times$ memory space and 1.9$\times$ inference speed. 
Our contributions can be summarized below:
\begin{itemize}
    \item We propose a training framework to smooth the representation space according to $k$NN knowledge.
    \item We devise an inject-and-refine training loop in our framework. Experiments show that  refreshing the datastore asynchronously matters.
    \item Our \method system achieves promising improvements and beats the state-of-the-art $k$NN-MT system.
\end{itemize}

%% file: Latex/02_background.tex
\section{Background}
This section briefly introduces the working process of $k$NN-MT and the architecture of adapter~\cite{bapna2019simple}.
For the latter, we will use it to improve the representation space in our framework.

\subsection{$k$NN-MT}
\label{sec:explicit}
Given an off-the-shelf NMT model $\mathcal{M}$ and training set $\mathcal{C}$, $k$NN-MT memorizes training examples explicitly with a \textit{key}-\textit{value} datastore $\mathcal{D}$ and use $\mathcal{D}$ to assist the NMT model during inference.

\noindent\paragraph{Memorize representations into datastore}~Specifically, we feed training example $(X, Y)$ in $\mathcal{C}$ into $\mathcal{M}$ in a teacher-forcing manner \cite{williams1989learning}.
At time step $t$, we record the contextualized representation\footnote{By default, the last decoder layer's output is used as the contextualized representation of the translation context ($X, Y_{<t}$).} ${h_t}$ as \textit{key} and the corresponding target token $y_t$ as \textit{value}. We then put the \textit{key-value} pair into the datastore.
In this way, the full datastore $\mathcal{D}$ can be created through a single forward pass over the training dataset $\mathcal{C}$:
\begin{equation}
\small
    \mathcal{D} = \{(h_t, y_t) ~|~ \forall y_t \in Y, (X, Y)\in \mathcal{C}\}
\end{equation}
where each datastore entry explicitly memorizes the mapping relationship between the representation $h_t$ and its target token $y_t$.

\begin{figure*}[ht]
    \centering
    \includegraphics[width=0.9\textwidth]{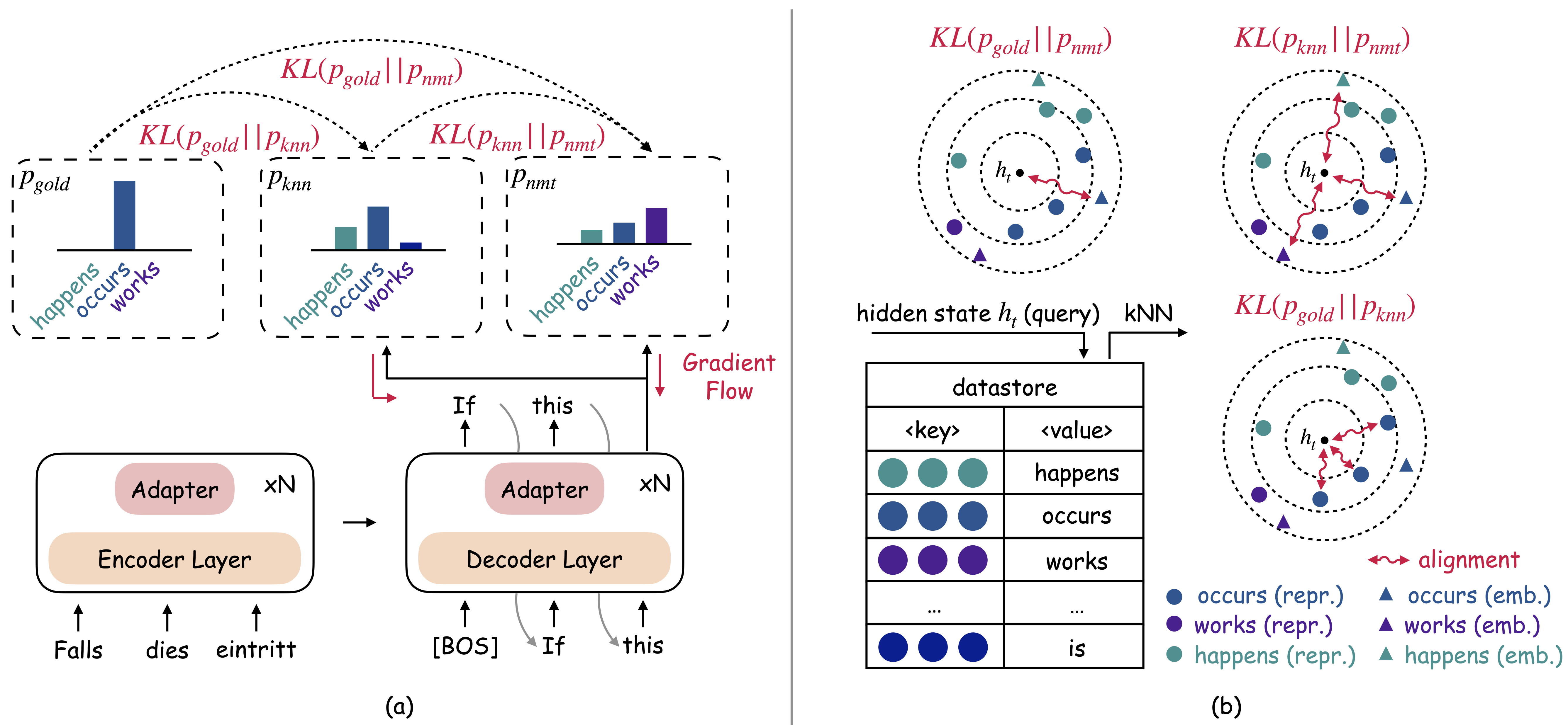}
    \caption{(a) Overview of our proposed training framework. (b) Illustration of how the three learning objectives encourage the refinement of the representation space. ``repr.'' and ``emb.'' denotes the contextualized representation and token embedding respectively. First, we align the contextualized representation and its target embedding to keep semantic meanings. We then align the contextualized representations of a target token and align the extract $k$NN  representations to address the issues of sparsely dispersing. }
    \label{fig:co_train}
\end{figure*}
\noindent\paragraph{Translate with memorized representations}~During inference, the contextualized representation of the test translation context $(X, Y_{<t})$ will be used to query the datastore for nearest neighbor representations and their corresponding target tokens $\mathcal{N}_k=\{(\hat{h}, \hat{y})\}_{1}^k$.
Then, the retrieved entries are converted to a distribution over the vocabulary:
\begin{equation}
\label{eq:knn}
\small
p_{\text{knn}}(y|X, Y_{<t}) \propto \sum_{(\hat{h}, \hat{y})\in \mathcal{N}_k} \mathbbm{1}(y=\hat{y}) e^{\frac{-d(h_t,\hat{h})}{T}}
\end{equation}
where $h_t$ denotes $h(X, Y_{<t})$ for short, 
$d$ measures Euclidean distance and $T$ is the temperature.

\subsection{Adapter}
\label{sec:implicit}
Previous research shows that adapter can be an efficient plug-and-play module for adapting an NMT model~\cite{bapna2019simple}.
In common, the adapter layer is inserted after each encoder and decoder layer of $\mathcal{M}$.
The architecture of the adapter layer is simple, which includes a feed-forward layer and a normalization layer.
Given the output vector $z\in \mathcal{R}^{d}$ of a specific encoder/decoder layer, the computation result of the adapter layer can be written as:
\begin{equation}
\small
    \widetilde{z} = W_2^T [W_1^T \cdot f(z)] + z
\end{equation}
where $f$ denotes layer-normalization, $W_1\in \mathcal{R}^{d\times d'}$, $W_2\in \mathcal{R}^{d' \times d}$ are two projection matrices.
$d'$ is the inner dimension of these two projections.
Bias term and activation function is omitted in the equation for clarity.
$\widetilde{z}$ is the output of the adapter layer.

%% file: Latex/03_method.tex
\section{Approach: INK}
This section introduces our training framework \method.  
The key idea of the proposed approach is to use $k$NN knowledge to smooth the representation space. The training process is built on a cycled loop: extracting $k$NN knowledge to adjust representations via a small adapter. The updated parameters are then used to refresh and refine the datastore to get new $k$NN knowledge. 
 We define three kinds of alignment loss to adjust representations, which are described in Section \ref{sec:basic}, Section \ref{sec:extraction}, and 
Section \ref{sec:refinement}. An illustration of the proposed framework is shown in Figure \ref{fig:co_train}.

\subsection{Align Contextualized Representations and Token Embeddings}
\label{sec:basic}
The basic way to optimize the adapter to minimize the KL divergence between the NMT system's prediction probability $p_{\text{nmt}}$ and the one-hot golden distribution $p_{\text{gold}}$:
\begin{equation}
\small
    \begin{split}
         \mathcal{L}^{a}_t & = D_{\text{KL}}[\ p_{\text{gold}}(y|X, Y_{<t})\parallel p_{\text{nmt}}(y|X, Y_{<t})\ ]  \nonumber \\
    & = -\log \frac{\sum_{(w, v)\in \mathcal{E}}\mathbbm{1}(v=y_t)\kappa(h_t, w)}{\sum_{(w,v)\in \mathcal{E}}\kappa(h_t, w)}
    \end{split}
\end{equation}
where $\mathcal{E}$ is the embedding matrix. $w$ and $v$ denote the token embedding and its corresponding token respectively. $h_t$ denotes the contextualized representation $h(X, Y_{<t})$. $y_t$ denotes the target token. $\kappa(h_t, w)=e^{h_t^T w}$.
Following the widely-accepted alignment-and-uniformity theory~\citep{DBLP:conf/icml/0001I20}, this learning objective aligns the contextualized representation $h_t$ with the tokens embedding of its corresponding target token.

\subsection{Align Contextualized Representations and $k$NN Token Embeddings}
\label{sec:extraction}
Previous research in $k$NN-MT has shown that the nearest neighbors in the representation space can produce better estimation via aggregating $k$NN neighbors~\cite{khandelwal2021nearest, zheng2021adaptive, yang2022nearest}.
Apart from the reference target token, the retrieval results provide some other reasonable translation candidates.
Taking the translation case in Figure \ref{fig:co_train} as an example, retrieval results provide three candidate words, where both ``happens'' and ``occurs'' are possible translations.
Compared with the basic one-hot supervision signal, the diverse $k$NN knowledge in the datastore can be beneficial for building a representation space with more expressive abilities.

Therefore, we extract $k$NN knowledge by using the contextualized representation $h_t$ to query the datastore for nearest neighbors $\mathcal{N}_k=\{(\hat{h}, \hat{y})\}_{1}^k$ (illustrated in Fig. \ref{fig:co_train}).
For more stable training, we reformulate the computation process of $k$NN distribution as kernel density estimation (KDE)~\cite{parzen1962estimation}.

\noindent\paragraph{Formulation}
\label{sec:formulation}
The general idea of KDE is to estimate the probability density of a point by referring to its neighborhood, which shares the same spirit with $k$NN-MT.
The computation of $k$NN distribution can be written as:
\begin{equation}
\label{eq:kde}
\small
p_{\text{knn}}(y|X,Y_{<t}) = \frac{\sum_{(\hat{h}, \hat{y})\in\mathcal{N}_{k}}\mathbbm{1}(y=\hat{y})\kappa(h_t,\hat{h})}{\sum_{(\hat{h}, \hat{y})\in\mathcal{N}_{k}} \kappa(h_t,\hat{h})}
\end{equation}
where $\kappa$ can be set as any kernel function.
Thus, Equation \ref{eq:knn} can be seen as a special case of Equation \ref{eq:kde} by setting $\kappa(\cdot, \cdot)=e^{-d(\cdot, \cdot)/T}$.

After extracting $k$NN knowledge, we use it to smooth the representation space by by minimizing the KL divergence between the $k$NN distribution $p_{\text{knn}}$ and NMT distribution $p_{\text{nmt}}$:
\begin{equation}
\small
    \begin{split}
        \mathcal{L}^{i}_t & = D_{\text{KL}}[\ p_{\text{knn}}(y|X, Y_{<t})\parallel p_{\text{nmt}}(y|X, Y_{<t})\ ] \nonumber \\
    & = -\sum_{\bar{y}\in \mathcal{Y}} p_{\text{knn}}(\bar{y})\cdot\log\frac{\sum_{(w, v)\in \mathcal{E}}\mathbbm{1}(v=\bar{y})\kappa(h_t, w)}{\sum_{(w,v)\in \mathcal{E}}\kappa(h_t, w)\cdot p_{\text{knn}}(\bar{y})}
    \end{split}
\end{equation}
    
where $\mathcal{Y}$ denotes identical tokens in nearest neighbors $\mathcal{N}_{k}$ and $p_{\text{knn}}(\bar{y})$ denotes $p_{\text{knn}}(y=\bar{y}|X, Y_{<t})$ for short.
$\mathcal{E}$ is the embedding matrix. $w$ and $v$ denote the token embedding and its corresponding token respectively. $h_t$ denotes $h(X, Y_{<t})$ for short. $\kappa$ is the kernel function. 
Following the widely-accepted alignment-and-uniformity theory~\citep{DBLP:conf/icml/0001I20}, this learning objective encourages $h_t$ to align with the embeddings of retrieved reasonable tokens, e.g., ``occurs'', ``happens''.

\subsection{Align Contextualized Representations of the Same Target Token}
\label{sec:refinement}
Although $k$NN knowledge could provide fruitful translation knowledge, it is also sometimes noisy \cite{zheng2021adaptive, jiang2022towards}.
For example, in Figure \ref{fig:co_train}, the retrieved word ``works'' is a wrong translation here.

To address this problem, we propose to adjust local representation distribution.  
Specifically, our solution is to optimize the $k$NN distribution towards the reference distribution by minimizing the KL divergence between the gold distribution $p_{\text{gold}}$ and $k$NN distribution $p_{\text{knn}}$.
Thanks to the new formulation (Eq. \ref{eq:kde}), we can choose kernel function here to achieve better stability for gradient optimization.
In the end, we find that exponential-cosine kernel works stably in our framework:
\begin{equation}
\kappa(h, h_t)=e^{\cos(h, h_t)}
\end{equation}
Therefore, the loss function can be written as:
\begin{equation}
\small
\label{eq:contrast}
    \begin{split}
         \mathcal{L}^{r}_{t} & = D_{\text{KL}}[\ p_{\text{gold}}(y|X, Y_{<t})\parallel p_{\text{knn}}(y|X, Y_{<t})\ ] \nonumber \\
    & = -\log \frac{\sum_{(\hat{h}, \hat{y})\in\mathcal{N}_k}\mathbbm{1}(\hat{y}=y_t)\kappa(h_t,\hat{h})}{\sum_{(\hat{h}, \hat{y})\in\mathcal{N}_{k}} \kappa(h_t,\hat{h})} \nonumber \\
    \end{split}
\end{equation}

where $\mathcal{N}_{k}$ is the retrieved k nearest neighbors. $\hat{h}$ and $\hat{y}$ denotes the neighbor representations and the corresponding target token. $h_t$ denotes $h(X, Y_{<t})$ for short.
Following the widely-accepted alignment-and-uniformity theory~\citep{DBLP:conf/icml/0001I20}, this learning objective aligns the contextualized representation of the same target token. With this goal, we can make the $k$NN knowledge less noisy in the next training loop by refreshing the datastore with the updated representations.

\subsection{Overall Training Procedure}
\label{sec:overall}
\noindent\paragraph{The combined learning objective}
To summarize, we adjust representation space via a small adapter with the combination of three alignment loss $\mathcal{L}_{t}^{a}$, $\mathcal{L}_{t}^{i}$, $\mathcal{L}_{t}^{r}$.
Given one batch of training examples $\mathcal{B}=\{(X,Y)\}$, the learning objective is minimizing the following loss:
\begin{equation}
\small
    \mathcal{L} = \frac{1}{|\mathcal{B}|}\sum_{(X,Y)\in\mathcal{B}}\sum_{t}(\mathcal{L}_{t}^{a} + \alpha\mathcal{L}^{i}_t + \beta\mathcal{L}^{r}_t)
\label{eq:objective}
\end{equation}
where $\alpha$, $\beta$ is the interpolation weight.
We notice that, in general, all three learning objective pull together closely related vectors and push apart less related vectors in the representation space, which has an interesting connection to contrastive learning \cite{lee2021contrastive, an2022cont} by sharing the similar goal.

\noindent\paragraph{Refresh datastore asynchronously}
In our training loop, once the parameters are updated, we refresh the datastore with the refined representation.
In practice, due to the computation cost, we refresh the datastore asynchronously at the end of each training epoch to strike a balance between efficiency and effectiveness
As the training reaches convergence, we drop the datastore and only use the optimized adapter to help the off-the-shelf NMT model for the target domain translation.

%% file: Latex/04_experiments.tex
\section{Experiments}
\subsection{Setting}
We introduce the general experiment setting in this section.
For fair comparison, we adopt the same setting as previous research of $k$NN-MT \cite{khandelwal2021nearest, zheng2021adaptive, jiang2022towards}, e.g., using the same benchmark datasets and NMT model. 
For training \method, we tune the weight $\alpha$ and $\beta$ among \{0.1, 0.2, 0.3\}. 
More implementation details are reported in the appendix.

\paragraph{Target Domain Data}
We use four benchmark German-English dataset (Medical, Law, IT, Koran) \cite{tiedemann2012parallel} and directly use the pre-processed data\footnote{\url{https://github.com/zhengxxn/adaptive-knn-mt}} released by \citet{zheng2021adaptive}. Statistics of four datasets are listed in Table \ref{tab:statistics}.

\begin{table}[ht]
    \centering
    \small
    \begin{tabular}{cccc}
    \toprule
    \textbf{Dataset} & \textbf{\# Train} & \textbf{\# Dev} & \textbf{\# Test} \\
    \midrule
    Medical & 248,099 & 2,000 & 2,000 \\
    Law     & 467,309 & 2,000 & 2,000 \\
    IT      & 222,927 & 2,000 & 2,000 \\
    Koran   & \hspace{0.13cm}17,982 & 2,000 & 2,000 \\
    \bottomrule
    \end{tabular}%}
    \caption{Statistics of four datasets. \#Train, \#Dev, \#Test represent the number of sentence pairs in training, development, and test sets, respectively.}
    \label{tab:statistics}
\end{table}

\paragraph{NMT Model}
We choose the winner model\footnote{\url{https://github.com/facebookresearch/fairseq/tree/main/examples/wmt19}} \cite{ng2019facebook} of WMT'19 German-English news translation task as the off-the-shelf NMT model for translation and datastore construction, which is based on the big Transformer architecture \cite{vaswani2017attention}.

\begingroup
\renewcommand{\arraystretch}{1.3} % Default value: 1
\begin{table*}[ht]
    \centering
    \footnotesize
    \resizebox{\textwidth}{!}
    {
    \begin{tabular}{lp{1.2cm}<{\centering}p{0.8cm}<{\centering}p{1.0cm}<{\centering}p{0.8cm}<{\centering}p{1.0cm}<{\centering}p{0.8cm}<{\centering}p{1.0cm}<{\centering}p{0.85cm}<{\centering}p{1.0cm}<{\centering}p{0.8cm}<{\centering}p{1.0cm}<{\centering}}
    \toprule
    \multirow{2}{*}{\hspace{0.8cm}\textbf{Systems}} & \multirow{2}{*}{\textbf{Mem.}} & \multicolumn{2}{c}{\textbf{Medical}} & \multicolumn{2}{c}{\textbf{Law}}    & \multicolumn{2}{c}{\textbf{IT}}     & \multicolumn{2}{c}{\textbf{Koran}} & \multicolumn{2}{c}{\textbf{Avg.}} \\
                           & & COMET & BLEU  & COMET & BLEU & COMET & BLEU & COMET & BLEU & COMET & BLEU \\
    \midrule
    Off-the-shelf NMT & - & 46.87 & 40.00 & 57.52 & 45.47 & 39.22 & 38.39 & -1.32  & 16.26 & 35.57 & 35.03 \\
    $k$NN-KD          & - & 56.20 & 56.37 & 68.60 & 60.65 & -1.57 & 1.48 & -13.05 & 19.60 & 27.55 & 34.53     \\
    \midrule
    \multicolumn{12}{c}{\textit{NMT + Datastore Augmentation}} \\
    \midrule
    V-$k$NN & $\times \textrm{1.7}$ & 53.46 & 54.27 & 66.03 & 61.34 & 51.72 & 45.56 & 0.73 & 20.61 & 42.98 & 45.45 \\
    A-$k$NN & $\times \textrm{1.7}$ & 57.45 & 56.21 & 69.59 & 63.13 & 56.89 & 47.37 & 4.68 & 20.44 & 47.15 & 46.79\\
    R-$k$NN$^\dagger$ & $\times \textrm{1.7}$ & 58.05 & 54.16 & 69.10 & 60.90 & 54.60 & 45.61 & 3.99 & 20.04 & 46.44 & 45.18 \\
    R-$k$NN  & $\times \textrm{43.8}$ & 57.70 & 57.12 & 70.10 & \textbf{63.74} & 57.65 & 48.50 & 5.28 & 20.81 & 47.68 & 47.54 \\
    \midrule
    \multicolumn{12}{c}{\textit{NMT + Representation Refinement}} \\
    \midrule
    Adapter & $\times \textrm{1.0}$ & 60.14 & 56.88 & 70.87 & 60.64 & 66.86 & 48.21 & 4.23  & 21.68 & 50.53 & 46.85 \\
    \textbf{\method} (ours) & $\times \textrm{1.0}$ & \textbf{61.64}$^{*}$ & \hspace{0.13cm}\textbf{57.75}$^{*}$ & \textbf{71.13} & \hspace{0.13cm}61.90$^{*}$ & \textbf{68.45}$^{*}$ & \hspace{0.13cm}\textbf{49.12}$^{*}$ & \hspace{0.13cm}\textbf{8.84}$^{*}$ & \hspace{0.13cm}\textbf{23.06}$^{*}$ & \textbf{52.52} & \textbf{47.85}\\
    \bottomrule
    \end{tabular}
    }
    \caption{Results on four datasets. ``Mem.'' stands for the added memory. ``COMET'' and ``BLEU'' are two metrics for evaluating translation performance. Scores shown in bold denote the highest performance among different systems. \method achieves better performance than the state-of-the-art $k$NN-MT system, i.e., R-$k$NN, with the least memory space. \method also outperforms the fine-tuned adapter baseline by a large margin. The annotation ``*'' indicates that the improvement is siginificant ($p<0.1$). R-$k$NN$^\dagger$ denotes the situation where the \textit{key} file of R-$k$NN is removed, and the approximate distance is used for inference. We can see that the state-of-the-art $k$NN-MT system still relies on the \textit{key} file to maintain a high level of translation performance.}
    \label{tab:main}
\end{table*}
\endgroup

\paragraph{Baselines}
For comparison, we consider three $k$NN-MT systems, which use datastore in different fashions. 
We report the translation performance of the adapter baseline to show the effectiveness of our training framework.
Besides, we report the translation performance of $k$NN-KD, which is another work using $k$NN knowledge to help NMT.
\begin{itemize}[itemsep=1pt]
    \item \textbf{V-$k$NN} \cite{khandelwal2021nearest}, the vanilla version of $k$-nearest-neighbor machine translation.
    \item \textbf{A-$k$NN} \cite{zheng2021adaptive}, an advanced variants of $k$NN-MT, which dynamically decides the usage of retrieval results and achieve more stable performance.
    \item \textbf{R-$k$NN} \cite{jiang2022towards}, the state-of-the-art $k$NN-MT variant, which dynamically calibrates $k$NN distribution and control more hyperparameters, e.g. temperature, interpolation weight.
    \item \textbf{Adapter} \cite{bapna2019simple}, adjusting representation by simply align contextualized representation and token embeddings.
    \item \textbf{$k$NN-KD} \cite{yang2022nearest}, aiming at from-scratch train a NMT model by distilling $k$NN knowledge into it.
\end{itemize}

\paragraph{Metric}
To evaluate translation performance, we use the following two metrics:
\begin{itemize}[itemsep=1pt]
    \item \textbf{BLEU} \cite{papineni2002bleu}, the standard evaluation metric for machine translation. We report case-sensitive detokenized \textit{sacrebleu}\footnote{\url{https://github.com/mjpost/sacrebleu}}.
    \item \textbf{COMET} \cite{rei2020comet}, a recently proposed metric, which has stronger correlation with human judgement. We report COMET score computed by publicly available \textit{wmt20-comet-da}\footnote{\url{https://github.com/Unbabel/COMET}} model.
\end{itemize} 

\paragraph{Approximate Nearest Neighbor Search} 
We follow previous $k$NN-MT studies and use Faiss\footnote{\url{https://github.com/facebookresearch/faiss}} index \cite{johnson2019billion} to represent the datastore and accelerate nearest neighbors search.
Basically, the \textit{key} file can be removed to save memory space once the index is built.
But, it is an exception that R-$k$NN relies on the \textit{key} file to re-compute accurate distance between query representation and retrieved representations.

\subsection{Main Results}
\label{sec:main}
We conduct experiments to explore the following questions to better understand the effectiveness of our proposed framework and relationship between two ways of smoothing predictions:
\begin{itemize}[itemsep=1pt]
    \item \textbf{RQ1}: \textit{Can we smooth the representation space via small adapter and drop datastore aside during inference?}
    \item \textbf{RQ2}: \textit{How much improvement can be brought by using $k$NN knowledge to adjust the representation distribution?}
    \item \textbf{RQ3}: \textit{Will together using adapter and datastore  bring further improvement?}
\end{itemize}

\noindent\paragraph{\method system achieves the best performance by smoothing the representation space}
Table \ref{tab:main} presents the comparison results of different systems. 
Due to the poor quality of representation space, the off-the-shelf NMT model does not perform well.
The performance of $k$NN-KD is unstable, e.g., it performs poorly on IT dataset.
$k$NN-MT systems generate more accurate translation. 
Among them, R-$k$NN achieves the best performance, which is consistent with previous observation \cite{jiang2022towards}.
Our \method system achieves the best translation performance with the least memory space.
Compared with the strongest $k$NN-MT system, i.e. R-$k$NN, INK achieves better performance on three out of four domains (Medical, IT, Koran).
In average, \method outperforms R-$k$NN with an improvement of 4.84 COMET and 0.31 BLEU while occupying 0.02$\times$ memory space.

\begin{figure*}[ht]
    \centering
    \includegraphics[width=\textwidth]{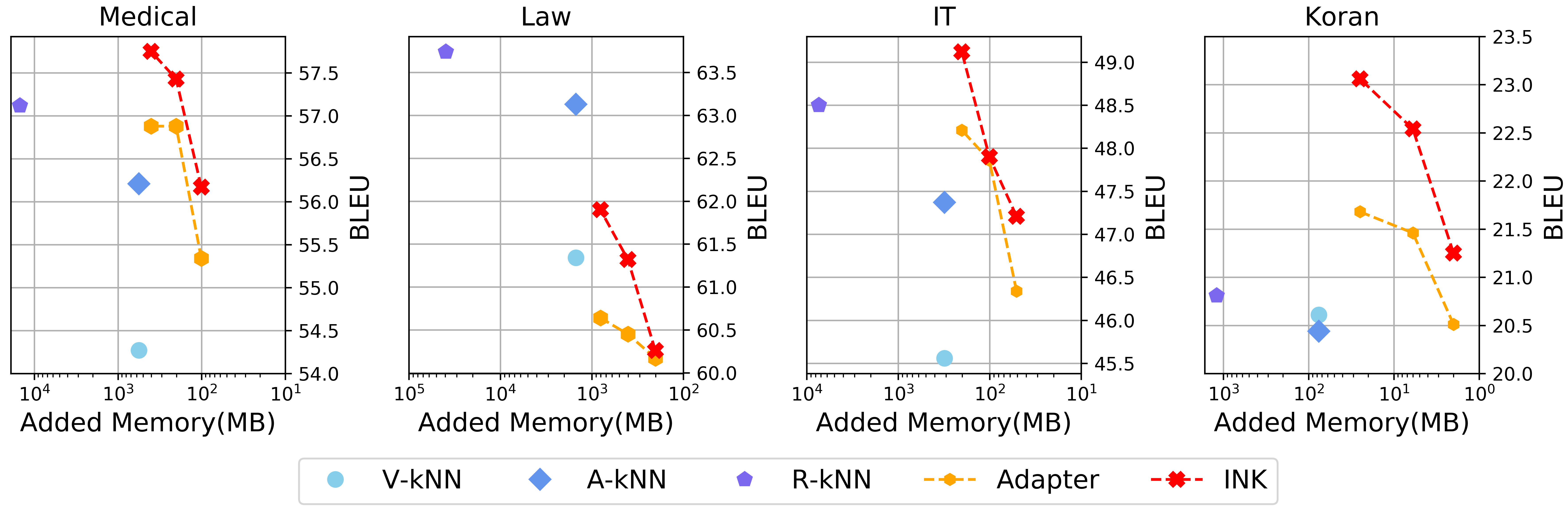}
    \caption{Comparison on added memory and BLEU scores on four datasets. Generally, representation-refined \method system achieves better performance than $k$NN-MT systems with less memory. Compared with adapter baseline, \method brings large improvement of the BLEU score in most cases.}
    \label{fig:memory}
\end{figure*}

\begin{figure}[ht]
    \centering
    \includegraphics[width=0.45\textwidth]{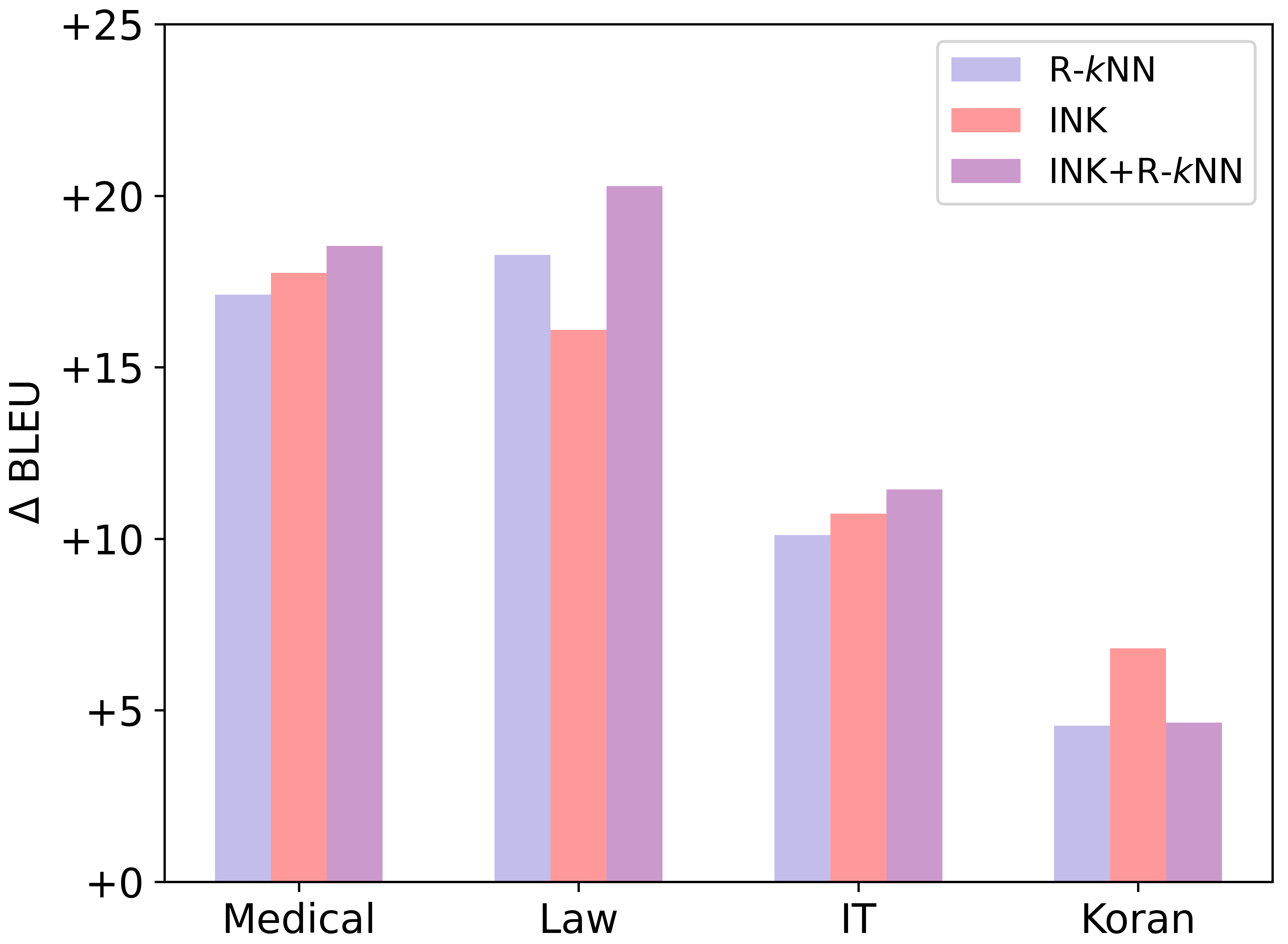}
    \caption{BLEU scores improvement brought by applying three different systems on four datasets. Using \method and R-$k$NN together brings further improvement on Medical, Law and IT.}
    \label{fig:hybrid}
\end{figure}

\noindent\paragraph{Representation refinement according to $k$NN knowledge brings large performance improvement}
\label{sec:size}
In Table \ref{tab:main}, compared with the adapter baseline that simply align the contextualized representations and word embeddings, \method outperforms it by 1.99 COMET and 1.00 BLEU in average, which demonstrates the effectiveness of adjusting representation distribution with $k$NN knowledge.
To better show the effect of \method framework, we use adapters of different sizes to refine the representation space.
Figure \ref{fig:memory} shows the BLEU scores and added memory of different systems on four datasets.
We can see that representation-refined system occupies much less memory than the datastore-enhanced system.
In general, \method systems locates on the top-right of each figure, which means that \method achieves higher BLEU scores with less memory space.
In most cases, \method outperforms adapter with a large margin, which demonstrates the superiority of our training framework.

\begin{table*}[ht]
    \footnotesize
    \centering
    \begin{tabular}{cp{1.2cm}<{\centering}p{1.3cm}<{\centering}p{1.3cm}<{\centering}p{1.3cm}<{\centering}p{1.3cm}<{\centering}p{1.3cm}<{\centering}p{1.3cm}<{\centering}}
    \toprule
    Mean $k$NN Acc (\%) & Systems & [0, 1k) & [1k, 5k) & [5k, 10k) & [10k, 20k) & [20k,~30k) & [30k, 42k) \\
    \midrule
    \multirow{2}{*}{$k$=8}  & NMT & 77.75 & 73.25 & 71.88 & 66.00 & 64.38 & 51.13 \\
                        & \method & \textbf{84.25} & \textbf{79.00} & \textbf{77.63} & \textbf{72.25} & \textbf{70.50} & \textbf{84.13} \\
    \midrule
    \multirow{2}{*}{$k$=16} & NMT & 76.25 & 70.88 & 69.13 & 63.19 & 61.31 & 34.06 \\
                        & \method & \textbf{83.81} & \textbf{77.31} & \textbf{75.75} & \textbf{70.00} & \textbf{67.88} & \textbf{79.50} \\
    \midrule
    \multirow{2}{*}{$k$=32} & NMT & 74.59 & 68.06 & 66.25 & 60.19 & 57.31 & 30.13 \\
                        & \method & \textbf{83.41} & \textbf{75.41} & \textbf{73.50} & \textbf{67.44} & \textbf{54.84} & \textbf{57.09} \\
    \midrule
    \multirow{2}{*}{$k$=64} & NMT & 72.97 & 64.89 & 62.97 & 56.67 & 52.22 & 28.13 \\
                        & \method & \textbf{83.20} & \textbf{73.16} & \textbf{70.80} & \textbf{64.31} & \textbf{60.38} & \textbf{43.05} \\
    \bottomrule
    \end{tabular}
    \caption{The quality of different systems' representation space. We use mean $k$NN accuracy as the evaluate metric and evaluate representations correspond to different tokens (the higher the token id, the lower the token frequency.). Bold text denotes the higher score in the two system. \method consistently improves the representation distribution, especially for low-frequency tokens.}
    \label{tab:acc}
\end{table*}

\noindent\paragraph{Jointly applying adapter and datastore can further smooth predictions}
Given the fact that both \method and datastore can smooth predictions, we take a step further and explore to use them together as a hybrid approach.
Specifically, on top of our \method system, we follow the fashion of R-$k$NN to use an additional datastore to assist it during inference.
Experiment results are shown in Figure \ref{fig:hybrid}.
On three out of four datasets, we can observe further improvements over \method.
On the Law dataset, the performance improvement even reaches 4.19 BLEU.
On the Medical and IT dataset, the performance improvement is 0.71 BLEU and 0.79 BLEU respectively.
Such phenomenon indicates that the representation space of the NMT model is not fully refined by the adapter.
If a more effective framework can be designed, the benefit of smoothing representation space will be further revealed.
The results on the Koran dataset is an exception here.
We suggest that it is because of the sparse training data, which makes it difficult to accurately estimate $k$NN distribution during inference.

%% file: Latex/05_analysis.tex
\section{Analysis and Discussion}
We conduce more analysis in this section to better understand our \method system.

\noindent\paragraph{\method greatly refines the representation space of the NMT model}
Inspired by \citet{li2022better}, we evaluate the quality of the representation space by computing mean $k$NN accuracy, which measures the ratio of k-nearest representations sharing the same target token with the query representation.
Ideally, all of the representations in a neighborhood should share the same target token.
Here, we use the contextualized representations from the unseen development set as the query.
For each query, the nearest representations from the training set will be checked.
Table \ref{tab:acc} shows the evaluation results on medical dataset.
\method achieves higher accuracy than the NMT model consistently.
For low frequency tokens, the representation quality gap is especially large.
\begin{table}[ht]
    \footnotesize
    \centering
    \begin{tabular}{lcc}
    \toprule
    Systems              &  BLEU  & $\Delta$\\
    \toprule
    \method w/o \textit{datastore refresh}    &  56.95 & -0.80 \\
    \method w/o $\mathcal{L}^{r}_t$  &  57.25 & -0.50 \\
    \method w/o $\mathcal{L}^{i}_t$  &  57.26 & -0.49 \\
    \midrule
    \method                                   &  57.75 & - \\
    \bottomrule
    \end{tabular}
    \caption{Ablation study for our \method framework on Medical dataset. All techniques introduced in \method are necessary. Asynchronously refreshing the datastore is important for smoothing representations.}
    \label{tab:ablation}
\end{table}

\noindent\paragraph{Ablation study}
To show the necessity of different proposed techniques in our \method framework, we conduct ablation study in this section.
In Table \ref{tab:ablation}, we can see that keeping the datastore frozen degenerates the translation performance most, which demonstrates the necessity of refreshing datastore asynchronously during training.
Removing either of the two alignment loss ($\mathcal{L}^{i}_t$ and $\mathcal{L}^{r}_t$) would cause the translation performance to decline, which validates their importance for adjusting the representation distribution.

\noindent\paragraph{\method enjoys faster inference speed}
After refining the representation space, our adapted system no longer need to querying datastore during inference.
We compare the inference speed~\footnote{We evaluate the inference speed on a single NVIDIA Titan-RTX.} of \method  and R-$k$NN.
Considering that decoding with large batch size is a more practical setting \cite{helcl2022non}, we evaluate their inference speed with increasing batch sizes.
To make our evaluation results more reliable, we repeat each experiment three times and report averaged inference speed.  
Table \ref{tab:speed} shows the results.
As the decoding batch size grows, the speed gap between the two adapted system becomes larger.
Our \method can achieve up to 1.9$\times$ speedup.
Besides, due to the fact that neural parameters allows highly parallelizable computation, the inference speed of \method may be further accelerated in the future with the support of non-autoregressive decoding \cite{qian2021glancing, bao2022textit}.

\begin{table}[htbp]
    \footnotesize
    \centering
    \begin{tabular}{lccc}
    \toprule
    Systems        & Batch=8 & Batch=32 & Batch=128 \\
    \midrule
    R-$k$NN       & 14.0 & 26.1 & 29.4 \\
    \method       & 19.9 & 46.4 & 55.1 \\
    \midrule
    \textit{Speedup}  & 1.4$\times$ & 1.8$\times$ & 1.9$\times$ \\
    \bottomrule
    \end{tabular}
    \caption{Inference speed (sents/s) of MT systems on Law dataset. Compared with R-$k$NN, \method enjoys up to 1.9$\times$ speedup on inference speed.}
    \label{tab:speed}
\end{table}

%% file: Latex/06_related_work.tex
\section{Related Work}
\noindent\paragraph{Nearest Neighbor Machine Translation}
$k$NN-MT presents a novel paradigm for enhancing the NMT system with a symbolic datastore.
However, $k$NN-MT has two major flaws: (1) querying the datastore at each decoding step is time consuming and the datastore occupies large space. (2) the noise representation in the datastore can not be easily updated, which causes the retrieval results to include noise.

Recently, a line of work focuses on optimizing system efficiency.
\citet{martins2022efficient} and \citet{wang2022efficient} propose to prune datastore entries and conduct dimension reduction to compress the datastore.
\citet{meng2022fast} propose to in-advance narrow down the search space with word-alignment to accelerate retrieval speed.
\citet{martins2022chunk} propose to retrieve a chunk of tokens at a time and conduct retrieval only at a few decoding steps with a heuristic rule.
However, according to their empirical results, the translation performance always declines after efficiency optimization.

To exclude noise in the retrieval results, \citet{zheng2021adaptive} propose to dynamically decide the usage of retrieved nearest neighbors with a meta-$k$ network. 
\citet{jiang2022towards} propose to dynamically calibrate the $k$NN distribution and control more hyperparameters in $k$NN-MT.
\citet{li2022better} propose to build datastore with more powerful pre-trained models, e.g. XLM-R \cite{conneau2020unsupervised}. 
However, all of this methods rely on a full datastore during inference.
When the training data becomes larger, the inference efficiency of these approaches will becomes worse.
Overall, it remains an open challenge to deploy a high-quality and efficient $k$NN-MT system.

\noindent\paragraph{Using $k$NN knowledge to build better NMT models}
As datastore stores a pile of helpful translation knowledge, recent research starts exploring to use $k$NN knowledge in the datastore to build a better NMT model.
As an initial attempt, \citet{yang2022nearest} try to from scratch train a better NMT model by distilling $k$NN knowledge into it.
Different from their work, we focus on smoothing the representation space of an off-the-shelf NMT model and enhancing its generalization ability via a small adapter.
Besides, in our devised inject-and-refine training loop we keep datastore being asynchronously updated, while they use a fixed datastore.

%% file: Latex/07_conclusion.tex
\section{Conclusion}
In this paper, we propose a novel training framework \method, to iteratively refine the representation space of the NMT model according to $k$NN knowledge.
In our framework, we devise a inject-and-refine training loop, where we adjust the representation distribution by aligning three kinds of representation and refresh the datastore asynchronously with the refined representations to update $k$NN knowledge.
Experiment results on four benchmark dataset shows that \method system achieves an average gain of 1.99 COMET and 1.0 BLEU.
Compared with the state-of-the-art $k$NN system (Robust $k$NN-MT), our \method also achieves better translation performance with 0.02$\times$ memory space and 1.9$\times$ inference speed up.

%% file: Latex/08_limitation.tex
\section{Limitation}
Despite promising results, we also observe that refreshing and querying the datastore during training is time-consuming. Our proposed training framework usually takes 3$\times$ $\sim$ 4$\times$ training time. In future work, we will explore methods to improve training efficiency. We include a training loop to dynamically use the latest datastore to inject knowledge into neural networks. However, we still find that the $k$NN knowledge still helps the inference even after our training loops, demonstrating that there still remains space to improve the effectiveness of knowledge injection.

\section*{Acknowledgement}
We would like to thank the anonymous reviewers for their insightful comments. Shujian Huang is the corresponding author. This work is supported by National Science Foundation of China (No. 62176120), the Liaoning Provincial Research Foundation for Basic Research (No. 2022-KF-26-02).

%% file: Latex/09_appendix.tex
\appendix
\section{Used Scientific Artifacts}
Below lists scientific artifacts that are used in our work. For the sake of ethic, our use of these artifacts is consistent with their intended use.
\begin{itemize} [itemsep=1pt]
    \item \textit{Fairseq (MIT-license)}, a sequence modeling toolkit that allows researchers and developers to train custom models for translation, summarization and other text generation tasks.
    \item \textit{Faiss (MIT-license)}, a library for approximate nearest neighbor search.
\end{itemize}

\section{Implementation Details}
We reproduce baseline systems with their released code.
We implement our system with \textit{fairseq} \cite{ott2019fairseq}. 
Adam is used as the optimizer and \textit{inverse sqrt} is used as the learning rate scheduler. 
We set 4k warm-up steps and a maximum learning rate as 5e-4. 
We set batch size as 4096 tokens.
All \method systems are trained on a single Tesla A100.
During inference, we set beam size as 4 and length penalty as 0.6.